\pdfoutput=1
\documentclass[11pt]{article}

\usepackage{acl}

\usepackage{times}
\usepackage{latexsym}
\usepackage[T1]{fontenc}
\usepackage[utf8]{inputenc}
\usepackage{latexsym}
\usepackage{graphicx}
\usepackage{booktabs}
\usepackage{todonotes}
\usepackage{comment}
\usepackage{multirow}
\usepackage{amsmath}

\usepackage{nicematrix}
\usepackage{multirow}

\usepackage{microtype}
\def\aclpaperid{756}

\title{Towards Debiasing Translation Artifacts}
\vspace{-3em}

\author{Koel Dutta Chowdhury\textsuperscript{{\normalfont 1}}, Rricha Jalota\textsuperscript{{\normalfont 1}}, \\ {\bf Cristina Espa\~{n}a-Bonet}\textsuperscript {{\normalfont 2}}{\bf,} \and {\bf Josef van Genabith}\textsuperscript{1,2}  \\  
  \textsuperscript{1}Saarland University, Saarland Informatics Campus, Germany \\
  \textsuperscript{2}German Research Center for Artificial Intelligence (DFKI) 
  \\
 {\tt koel.duttachowdhury@uni-saarland.de}\\
  {\tt rrja00001@stud.uni-saarland.de}\\ 
  {\tt \{cristinae, Josef.Van\_Genabith\}@dfki.de }\\
  }

\date{}

\begin{document}
\maketitle
\vspace{-10em}
\begin{abstract}

Cross-lingual natural language processing relies on translation, either by humans or machines, at different levels, from translating training data to translating test sets. However, compared to original texts in the same language, translations  possess distinct qualities referred to as translationese.
Previous research has shown that these translation artifacts influence the performance of a variety of cross-lingual tasks.
In this work, we propose a novel approach to reducing translationese by extending an established bias-removal technique. We use the Iterative Null-space Projection (INLP) algorithm, and show by measuring classiﬁcation accuracy before and after debiasing, that translationese is reduced at both sentence and word level.
We evaluate the utility of debiasing translationese on a natural language inference (NLI) task, and show that by reducing this bias, NLI accuracy improves. To the best of our knowledge, this is the first study to debias translationese as represented in latent embedding space.

\end{abstract}

\section{Introduction}
\label{s:intro}
“Translationese” refers to features in (professional human or machine) translated text that distinguishes it from non-translated, original text in the same language.
Carriers of translationese include lexical and word order choices that are influenced by the source language \cite{gellerstam:1986}, as well the use of more explicit and standardised constructions \cite{Baker1993CorpusLA} compared to original text. 
Translationese has a significant
impact in machine translation evaluation. \citet{toral-etal-2018-attaining} found that translating source sentences that are already the result of translation are easier to translate than original sentences.
Similarly, \citet{edunov-etal-2020-evaluation}
show that back-translation results in large BLEU scores when translationese is on the source side and original text is used as reference.
To avoid such artifacts, 
it is advised to use original source sentences for machine translation  evaluation~\cite{zhangToral:2019,graham-etal-2020-statistical}. \citet{rileyEtAl:2020}  train sentence-level classifiers to differentiate translationese from original target text, and then use this classifier to tag the training data for an NMT model to produce output that shows fewer translationese effects.
Translation-inherited artifacts have been shown to have significant impact on other tasks as well. 
For example, \citet{singh2019xlda} show that substituting segments of original training samples by their translations from another language improves performance on natural language inference (NLI) tasks. ~\citet{10.1162/tacl_a_00317} introduce a translation-free Question Answering dataset to avoid having inflated gains from translation artifacts in transfer-learning tasks.
\citet{artetxe-etal-2020-translation} show that cross-lingual models suffer from induced translation artifacts when evaluated on translated test sets.
These examples motivate the need for reducing translation artifacts. 

While a number of methods to remove or attenuate human-like biases (e.g., gender, race, etc.) in both static and contextualised word embeddings have recently been proposed \citep{bolukbasi2016man,gonen2019lipstick,dev2019attenuating,ravfogel2020null,liang-etal-2020-monolingual,zhou-etal-2021-challenges}, attenuating and eliminating a more implicit signal like translationese in embeddings has yet to be studied.  Translationese signals are complex and multi-faceted and, unlike e.g. gender and profanity, can in general not be captured in terms of simple lists of contrastive word pairs (woman-man, she-he, etc.), but rather manifest as a complex mix of morphological, lexical, syntactic and semantic phenomena.
Our study is a first attempt to directly debias translationese encoded as latent representations, based on the recently proposed Iterative Nullspace Projection (INLP) algorithm \cite{ravfogel2020null}. 

The main contributions of our work are as follows. $(i)$  We propose to reduce the bias induced by translation by extending the INLP approach to translationese debiasing at both the word and sentence level, with findings demonstrating that our debiasing approaches can effectively attenuate translationese biases in both static word embedding spaces as well as sentence representations based on contextualised embeddings.
$(ii)$ We use the INLP method with a number of neural sentence-level translationese classification architectures (fastText, mBERT, XLM), and propose two alternative methods for detecting explicit translationese bias in word embeddings, and find that  after debiasing, the models' performance on classifying translationese degrades to that of a random classifier.
$(iii)$ Finally, by integrating the proposed debiasing method within the NLI task, we show the effect of translation artifacts and their removal on the task.

\section{Debiasing Strategies}
\label{s:inlp}
Much previous research \cite{bolukbasi2016man,zhao-etal-2018-gender,dev2019attenuating, ravfogel2020null} focuses on eliminating bias in word embeddings.  However, existing models use lists of contrastive word pairs (e.g., woman-man, she-he) to detect, capture and mitigate specific biases (e.g., gender, profanity, etc.).
While translationese cannot, in general, be captured in simple lists of contrastive word pairs, we do have labeled data at sentence level: translated (translationese) and original sentences. We use this data and the INLP algorithm to directly mitigate translationese at sentence level (Section~\ref{s:sentence}).
Additionally, we explore INLP at word level to debias translationese in Section~\ref{s:word}. 
Only a few earlier studies \citep{chowdhury2020understanding, chowdhury2021graph} deal with translationese at the level of word embedding spaces leveraging distances between graph-based representations of original and translationese data. For debiasing word embedding spaces, we adapt an idea from \citet{gonen-etal-2020-simple} to extract lists of pairs of identical words and examine how their use differs in translationese and original data (rather than contrasting word pairs). If a word is used very differently in translated and original data, this is reflected in differences in original and translated word embedding spaces, and is evidence of translationese in the embeddings. 
Alternatively, we propose a simpler approach that builds on a joint embedding space where words are tagged according to their origin (translationese or original) and without any need for a word list.

The \textbf{Iterative Nullspace Projection algorithm} \citep{ravfogel2020null} focuses on removing linearly decipherable features from vector representations originally for gender bias mitigation. 
Given a set of labeled data with data points $X = x_1,..., x_n$ and task labels $Y = y_1,..., y_n$, we use a standard classification setup with a neural network and a simple classifier $\tau$ on top. 
 An encoder $h$ encodes $x_k$ into a representation vector $h(x_k)$. $\tau$ predicts $y_k$ based on $h(x_k)$, i.e., $y_k$ = $\tau(h(x_k))$. Let $T$ be the trait we want to mitigate, referred to as the \textit{protected attribute}.
The goal of the INLP method is to neutralise the ability of the classifier $\tau$ to linearly predict $T$ from $h$.
$\tau$ is parameterised by a matrix $W$ and trained to predict $T$ from $h$.  Using $W$, one can collapse the data onto its nullspace $N(W)$ with a projection matrix $P_{N(W)}$. This guarantees $W P _{N(W)} h(T) = 0$, i.e., the information used to classify $T$ is linearly removed.
By repeating this process $i$ times until no classifier achieves above-majority accuracy, INLP can neutralise all features that $W_i$
uses for predicting $T$ from $\tilde{h}$:
\begin{equation}
\small
\tilde{h} := P_{N(W_1)}P_{N(W_2)}...P_{N(W_i)}h
\end{equation}
Details of the implementation are the same as in \citet{ravfogel2020null}.
\footnote{Our code is available at \url{https://github.com/koeldc/Towards-Debiasing-Translation-Artifacts/}}
\section{Translationese in Sentence Embeddings}
\label{s:sentence}

In our work, we are interested in adapting INLP to study the impact of removing the translationese attributes $T$ from semantic representations, via a binary classification task. Specifically, the binary classifier learns to distinguish between original and translationese sentences. Therefore, in our setup, labels $Y$ correspond to original and translationese and act as protected attributes.

\textbf{Data.} We use the Europarl corpus annotated with translationese information from \citet{europarl-motra21}. We focus on three languages: English (En), German (De) and Spanish (Es). The corpus provides originals in the three languages (L1) and translations into these three languages that come from original texts in the other two (L2). We use the notation L1--L2 to refer to the different sets in Table~\ref{sent-class} and Table~\ref{word-class}. For example, in En-De, L2 refers to English text translated from German. For each corpus, there is an equal number of translated and original sentences: 42k for De–En, De–Es, En–De, En–Es, Es–De and Es–En.
We use 70\% of the sentences for training, 15\% for development and 15\% for testing.

\begin{table}
\small
\resizebox{1\columnwidth}{!}{%
\begin{tabular}{c ccccc cc}
\toprule
 &\textbf{fastText} &\textbf{mBERT} 
& \textbf{mBERT} & \textbf{XLM} & \textbf{de-} \\
    &\textbf{Text} & \texttt{CLS} &  \textbf{${pool}$} & \texttt{CLS} & \textbf{biased}  \\
\midrule
En-De  &0.64 & 0.73 & 0.79 & 0.71 &0.50  \\
En-Es  &0.71 & 0.78 & 0.83 & 0.77 &0.50  \\
De-En &0.68 & 0.78 & 0.84 & 0.77 &0.50 \\
De-Es  &0.69 & 0.79 & 0.86 & 0.79 &0.50  \\
Es-De  &0.71 & 0.77 & 0.85 & 0.77 &0.50  \\
Es-En &0.72 & 0.76 & 0.82 & 0.77 &0.50  \\
\bottomrule
\end{tabular}
}\caption{Sentence Embedding Classiﬁcation accuracy on original versus translationese using different models. After INLP debiasing, translationese classification reduces to random 50\% accuracy in all cases.} 
\label{sent-class}
\end{table}

\textbf{Classifier.} We use a logistic classifier on top of sentence embeddings $h$ obtained with 4 models without any additional fine-tuning to the translationese classification task.
$(i)$ fastText \cite{joulin2016fasttext}: we compute an average of all token vectors in the sentence. 
$(ii)$ mBERT$_\texttt{CLS}$ \cite{devlinEtAl:2019}: we use the \texttt{[CLS]} token in mBERT as sentence representation. 
$(iii)$ mBERT$_{pool}$ \cite{devlinEtAl:2019}: we use mean pooling of mBERT's contextualised word embeddings. 
 $(iv)$ XLM$_\texttt{CLS}$ \cite{conneau2019unsupervised}: we use the \texttt{[CLS]} token from XLM-RoBERTa.

\textbf{Results.}
The first four columns in Table~\ref{sent-class} summarise the translationese classification accuracy achieved by the four models. 
mBERT$_{pool}$ achieves the best performance for all languages, while fastText trails the pack.
The final column in Table~\ref{sent-class} shows that INLP is close to perfection in removing translationse signals for the linear classifiers, reducing accuracy to a random 50\%. 

\section{Translationese in Word Embeddings}
\label{s:word}
Unlike sentence-level debiasing, word-level debiasing needs a seed translationese direction to obtain a debiased space. This is challenging for translationese as unlike, e.g. gender and profanity, translationese cannot in general be captured in terms of simple contrastive word pairs.
In what follows, we introduce two approaches for debiasing translationese at the level of word embeddings.

 \textbf{Stepwise Aligned Space Projection.}
 \label{aligned}
In order to estimate the seed translationese direction, we derive a list of words ($G$) used differently in translationese $T$ and original $O$ data using the \emph{usage change} concept from \citet{gonen-etal-2020-simple}.
The same word used in different data sets (original and translated) is likely to have different neighbours in the two embedding spaces. We only use  words from the intersection of both vocabularies $O$ and $T$. We compute the score for context change across the embeddings $\cal{O}$ and $\cal{T}$ of
the two data sets by considering the size of the intersection of two sets where each word in a corpus is represented as its top-$k$ nearest neighbors (NN) in its embedding space:
\begin{equation}
\small
    score ^k (w) = -|{\rm NN}_{\cal{O}}^k(w)\cap {\rm NN}_{\cal{T}}^k(w)|
    \label{eq:score}
\end{equation}
where NN$_i^k(w)$ is the set of $k$-NN of word $w$ in embedding space $i$.
The smaller the size of the intersection, the more differently the word is used in the two data sets (and words with the smallest intersection can be seen as indicators of translationese).
Given $\cal{O}$ and $\cal{T}$, we collect a ranked  list of about 500 words with the smallest intersection as our translationese word list $G$. $G$ allows us to identify the seed translationese direction for INLP. In our experiments we only consider words attested at least 200 times in the data. Appendix~\ref{app:wordlists} shows the top 50 elements for all the word lists.
In our experiments, we use the translated and original parts of the data described in Section~\ref{s:sentence} to estimate the word embeddings $\mathcal{O}$ and $\mathcal{T}$ and use $k$=1000 nearest neighbours in Equation~\ref{eq:score}. Following \citet{gonen-etal-2020-simple}, a large value of $k$ results in large neighbors sets for each word in the two corpora, resulting in a more stable translationese wordlist $G$. 

Next, we create a joint word embedding space $\mathcal{J}$ from the concatenation of the translated and original data, $T$ and $O$. Since this joint space $\mathcal{J}$ includes both original and translationese signals, we then align the previously unrelated $\mathcal{O}$ and $\mathcal{T}$ spaces to this embedding space $\mathcal{J}$, using VecMap~\cite{artetxe2018acl},
producing aligned spaces $\mathcal{\tilde{O}}$ and $\mathcal{\tilde{T}}$, and resulting in an extended single embedding space where $\mathcal{T}$ and $\mathcal{O}$ are aligned to $\mathcal{J}$. Next, we compute the translationese direction $v$
of the same word $w$ in the two embeddings spaces, $\mathcal{\tilde{O}}$ and $\mathcal{\tilde{T}}$, using,
\begin{equation}
\small
v:=  \mathcal{\tilde{T}}[w] - \mathcal{\tilde{O}}[w], \forall w \in G 
\label{direction}
\end{equation}
Finally, we compute the similarity of words in $\mathcal{J}$ along the directions $v$ and $-v$, to divide them into two subspaces
, translationese and non-translationese, respectively. 
Using Equation \ref{direction}, we initialise the INLP algorithm in two ways: $(i)$ \textbf{INLP.single}: with a direction vector created from the difference between two aligned spaces for the highest ranked word in $G$, and $(ii)$ \textbf{INLP.avg}: by averaging the differences of all words in $G$.

 \begin{table}
\small
\resizebox{1\columnwidth}{!}{%
\begin{tabular}{ccccc}
\toprule
\small
& {\textbf{Direct Joint}} &\multicolumn{2}{c}{{\textbf{Stepwise Aligned}}} &\textbf{de-}\\
&  & INLP.Single &INLP.Avg   &\textbf{biased} \\
\midrule
En-De &0.98  &0.99 &1.00&0.50\\
En-Es & 0.92 &1.00 &1.00& 0.50\\
De-En &0.96   &1.00 &0.99 &0.50 \\
De-Es &0.91   &1.00 & 1.00 &0.50  \\
Es-De &0.93  &0.99&0.99&0.50\\
Es-En &0.95  &1.00&1.00&0.50 \\
\bottomrule
\end{tabular}
}\caption{Classiﬁcation accuracy on original versus translationese with word embeddings using our two approaches, before and after debiasing with INLP. } 
\label{word-class}
\end{table}

 \textbf{Direct Joint Space Projection.}
 \label{single}
 In this approach, we directly build the embeddings of a specific word $w$ from $\mathcal{O}$ and $\mathcal{T}$ into the same space $\mathcal{J'}$ by annotating $w$ as either $w_o$ and $w_t$ on surface in the $O$ and $T$ data. In this way we can easily track and distinguish the two embeddings of the same word $w$ coming from $O$ and $T$ data in the same embedding space $\mathcal{J'}$ resulting from the simple concatenation of two datasets. This eliminates the need to compute the translationese direction vector $v$ to group the subspaces and the complexity of maintaining and aligning $\mathcal{O}$, $\mathcal{T}$, $\mathcal{J}$, $\mathcal{\tilde{O}}$ and $\mathcal{\tilde{T}}$  spaces. Figure~\ref{dendogram} in Appendix~\ref{image} shows the t-SNE projection \citep{van2008visualizing} of the tagged tokens before and after debiasing.
 
\textbf{Results.}
We compare the performance before and after debiasing using our two word-level debiasing methods in Table \ref{word-class}. As expected, debiasing reduces classification accuracy for all language pairs from $\sim$100\% to $\sim$50\% for both methods.

\textbf{Translationese Word Lists.} The content of our extracted translationese word lists depends on the language (see Appendix~\ref{app:wordlists}). For Es, punctuation is clearly used differently in originals and translations into Es. For De, pronominal adverbs play an important role, especially when translations come from Es. For En, there is no clear trend but, interestingly, only one word in the top-50 list (\emph{indeed}) overlaps with the words with a highest difference in frequency of usage in the original and translationese corpus as analysed for Europarl in \citet{koppel-ordan-2011-translationese}. The number of times and the context where a word appears may reflect two different aspects of translationese.
\section{Application to NLI}
\label{s:nli}

In order to investigate the impact of removing translationese artifacts from the translated data we analyse its impact on the NLI task, where machine translation is used.
NLI predicts the relationship between two sentences, premise and hypothesis, and classifies it into one of the three categories ---entailment, contradiction, or neutral.
Recently, \citet{artetxe-etal-2020-translation} showed that, in the existing NLI datasets, there exists a significant lexical overlap between the premise and the hypothesis, which is utilised by neural NLI models to make predictions with high accuracies. However, when the premise and hypothesis are paraphrased independently using translation and back-translation, lexical overlap is reduced, negatively impacting the performance of the models.

Below we test whether and if so, to which extent our INLP-based translationese debiasing approach can avoid the observed performance loss in the back-translated NLI task. We generate Back-Translated (BT) NLI dataset by independently translating premise and hypothesis from the Original SNLI data.
Then, we train two NLI models, one on Original and one on BT data. To reduce the translationese artifacts resulting from back-translation, we apply the sentence and word embedding debiasing strategies as described in Sections \ref{s:sentence} and \ref{s:word} over the embeddings generated from BT data, and use the resulting debiased-BT embeddings to train a third debiased NLI model in Table~\ref{tab:snli}.

Finally, we consider two scenarios to evaluate translationese debiasing - $(i)$  Symmetric \texttt{[Sym]}: where Original is tested with original test set, and BT and debiased NLI models are tested with BT test data and $(ii)$ Asymmetric \texttt{[Asym]}: where all these models are tested with original test data. We use $(i)$
to see whether debiasing the model trained on translated train-test can bring its performance closer to that of model trained on original train-test data. In $(ii)$, we examine the asymmetry between original test data and BT-NLI training data, and whether our translationese debiasing of BT training data can offset this asymmetry and improve NLI performance. Table~\ref{tab:snli} shows the classification accuracies with respect to $(i)$ and $(ii)$.  
\begin{table}[t]
    \centering
\resizebox{0.95\columnwidth}{!}{%
    \begin{tabular}{cc ccc}
\toprule 
 \textbf{Approach}  & \textbf{Test} & \multicolumn{3}{c}{\textbf{SNLI Model}}\\
  &   \textbf{Data}   & \textbf{Original} & \textbf{Back-translated} &\textbf{debiased} \\
       \midrule

          \multirow{2}{*}{Word-Joint}  &  \texttt{Sym}
     &\multirow{2}{*}{{67.2$\pm$0.1}}& 64.1$\pm$0.2 & 64.7$\pm$0.1 \\
       &       \texttt{Asym} & &64.4$\pm$0.2 &65.1$\pm$0.1\\
       \midrule
  \multirow{2}{*}{Word-Aligned}&  \texttt{Sym} &\multirow{2}{*}{67.8$\pm$0.1} & 64.3$\pm$0.2  & 64.8$\pm$0.2\\
&    \texttt{Asym}& &64.9$\pm$0.2& 65.3$\pm$0.1 \\
   \midrule
  \multirow{2}{*}{Sentence} &\texttt{Sym} & \multirow{2}{*}{65.6$\pm$0.0}&  58.4$\pm$0.0 & 60.1$\pm$0.0\\    
    &  \texttt{Asym} &&63.5$\pm$0.0 & 64.9$\pm$0.0\\ 
         \bottomrule
    \end{tabular}}
    \caption{Test set accuracies (3 runs mean) of our three NLI models in Symmetric and Asymmetric settings.}
    \label{tab:snli}
\end{table}

\textbf{Data.} The large-scale SNLI dataset (Stanford Natural Language Inference) \cite{bowman2015large} contains 570k sentence pairs in the training set manually classified as entailment, contradiction, or neutral. We use a subset of 10\% of the training data for our experiments. The development and test data are used as in the original SNLI splits (each containing 10k examples). 
We generate a back-translated variant of the training and test data using German (De) as the pivot language. For translation, we use the pre-trained models of Facebook-FAIR's WMT-19 news translation task submission \cite{ng-etal-2019-facebook}. 

\textbf{Models.}
We train three different NLI models, each for original, back-translated and debiased versions of back-translated embeddings.
For the word-level setup, we use a single hidden BiLSTM layer followed by a standard feedforward output layer on top of frozen fastText word embeddings for the 3-class NLI classification. 
The computation of word embeddings and debiasing follows the setup described in Section~\ref{s:word}.
For sentence-level debiasing, we use the BERT$_{pool}$ method explained in Section~\ref{s:sentence} with a linear SVC on top to predict the labels.

\textbf{Results.}
Table~\ref{tab:snli} shows results consistent with \citet{artetxe-etal-2020-translation}, in that models trained on Original data outperform models trained on BT data in both \texttt{Sym} and \texttt{Asym} scenarios. 
Table~\ref{tab:snli} also shows that, after translationese debiasing, classification accuracy on SNLI-debiased improves modestly for all models, with only a minor improvement at word-level, and larger improvement at the sentence level.
 Overall this may be due to the fact that translationese is a combination of lexical and syntactic phenomena that is better captured at sentence-level. Results in Table~\ref{tab:snli} suggest that debiasing translation artifacts helps in reducing the asymmetry between translated train and original test set. Therefore, rather than translating the entire test set to match the training set in transfer-learning tasks, debiasing the training set for translation artifacts is a promising direction for future work. Finally, for a complex task such as translationese debiasing, linear intervention alone may not be sufficient. As a result, non-linear guarding approaches need to be investigated further.
\section{Conclusion}
\label{s:conclusions}
In this work, we remove translationese artifacts by extending the debiasing INLP approach at both word and sentence level. To the best of our knowledge, this is the first paper that attempts at debiasing sentence and word embeddings for translationese.
We introduce two techniques for debiasing translationese at the word level: one (Stepwise Aligned Subspaces) is akin to the subspace construction approach of gender-debiasing proposed by \citet{bolukbasi2016man} and \citet{ravfogel2020null}, the second (Direct Joint Subspace) is a simplified approach that operates directly on the joint space without the use of a separate translationese word list and multiple independently computed and subsequently aligned subspaces. Our word-based debiasing study provides a systematic view of translationese biases contained in static embeddings.
We also explore translationese debiasing at sentence level embeddings computed from contextualised word embeddings. As expected, the INLP-based linear translationese debiasing results on static word embeddings are as “perfect” as our sentence level results, reducing the performance of a linear translationese classifier on the debiased data to chance, demonstrating that our debiasing strategies effectively attenuate translationese signals in both these spaces.

Further, we evaluate the effects of debiasing translation artifacts on a standard NLI task in two settings. Even though we achieve perfect performance for the translationese classification-based debiasing task with INLP, this translates into just modest improvements resulting from INLP-based debiasing translationese in neural machine translation in an NLI task, with slightly better results for sentence than word debiasing. This demonstrates that our debiasing approach is effective in reducing translation artifacts but that there is more to translationese than is visibe to a linear classifier.

Finally, we acknowledge that while this study is the first to debias translationese encoded as latent representation in (word and sentence) embedding space, the effect of this on the actual surface form of the generated output is not investigated. We hope to account for this in future work.

\section{Acknowledgments}

We would like to thank Saadullah Amin for his
helpful feedback. Funded by the Deutsche Forschungsgemeinschaft (DFG, German Research Foundation) – SFB 1102 Information Density and Linguistic Encoding.

\bibliography{acl2021}
\bibliographystyle{acl_natbib}
\newpage
\appendix

\section{Appendix}
\subsection{Experimental Setup and Hyperparameters}

Each run in Table~\ref{sent-class} took around 1.5 hours on a GTX1080Ti GPU. Each classification and debiasing step for SNLI sentence-level experiment in Table~\ref{tab:snli} took approximately 2 hours on V100-32GB GPU. Other hyperparameter settings are shown in the Table \ref{tab:setup}.

\begin{table}[h]
\centering
\small
\begin{tabular}{p{3cm}p{4cm}}
\toprule
Model & Hyperparameters \\
\midrule
fastText & minCount=5, dim=300 \\
Logistic Regression & warm\_start = True, penalty = 'l2', verbose=5, solver="saga", random\_state=23, max\_iter=7 \\
BiLSTM & hid\_dim=300, dropout=0.2, batch\_size=32, Adam optimizer with lr=0.0001, epochs=15\\
SNLI-sentence & nclfs=45, max\_iter =1500 \\
SNLI-Aligned (word) & nclfs=34 INLP.single,word=human \\
SNLI-Joint (word) & nclfs=35 \\
\bottomrule
\end{tabular}
\caption{Hyperparameter settings.}
\label{tab:setup}
\end{table}

\subsection{Debiased Word Representations}
\subsubsection{Word Analogy Tests}
\label{analogy}
To verify that debiasing does not hurt the quality of the word representations, we estimate the performance of the original and debiased embeddings on the word analogy task using the MultiSIMLEX benchmark \citep{vulic2020multi}. 
As MultiSIMLEX does not cover German, we use German-Simlex from \citet{leviant2015separated}. After debiasing, Spearman’s $\rho$ correlation coefficients show negligible decreases of 0.02 on En-De and En-Es, 0.01 on Es-En and Es-De, 0.3 on De-En and an increase of 0.01 for De-Es.

\begin{table}[b]
    \centering
\resizebox{0.95\columnwidth}{!}{%
    \begin{tabular}{cccccc}
    \toprule
       De-En  &  De-Es  &  En-De  &  En-Es  &  Es-De  &  Es-En      \\
    \midrule
    466 & 510 &429 &483 &547 &504 \\
     \bottomrule
    \end{tabular}}
    \caption{Size of translationese word lists created with the usage change algorithm \citep{gonen-etal-2020-simple}.}
    \label{tab:wordlist-stats}
\end{table}

\subsubsection{Visualisation}
\label{image}
\begin{figure}[h]
\centering
    \includegraphics[width=0.89\linewidth]{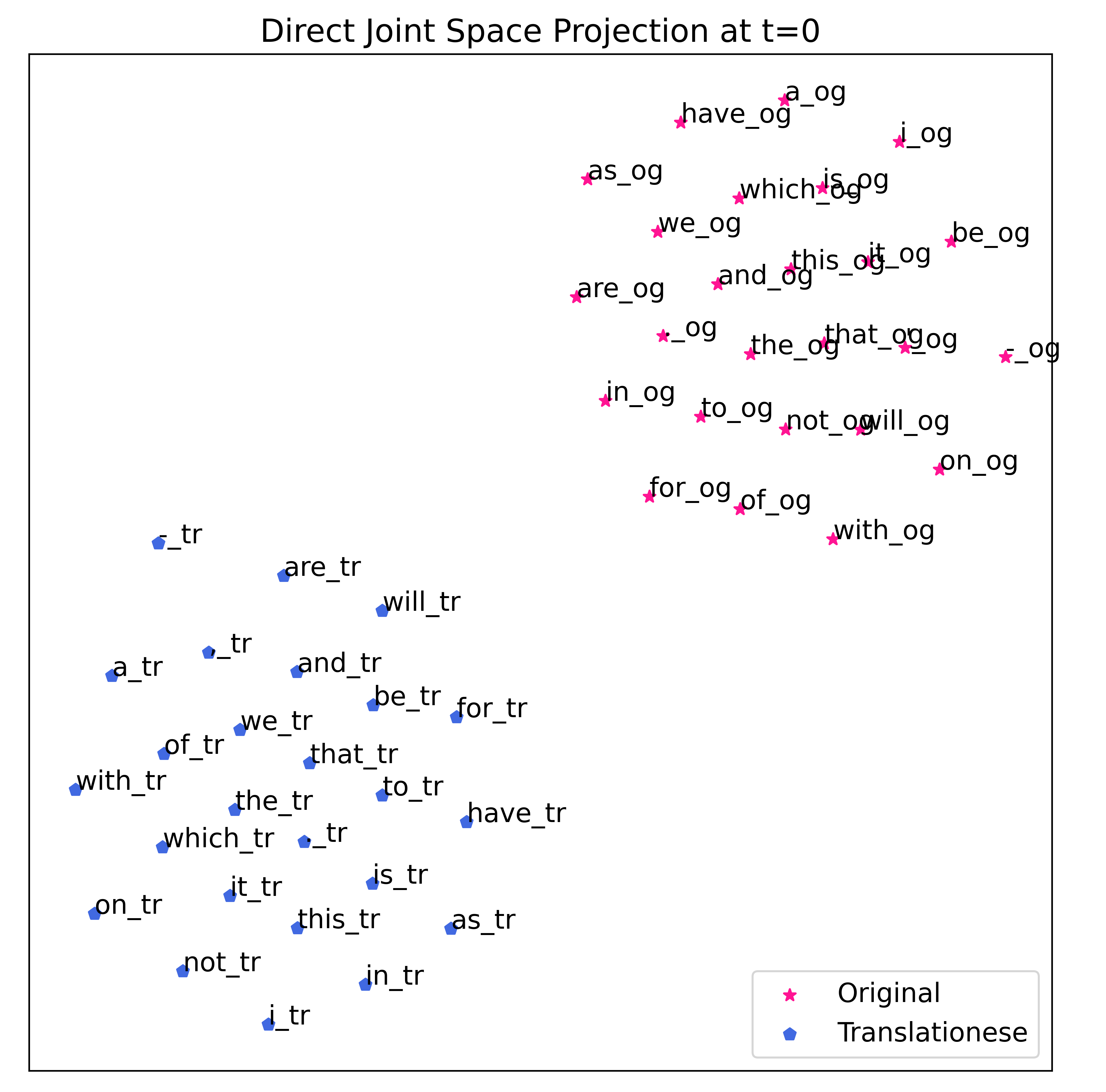}\\
  \includegraphics[width=0.85\linewidth]{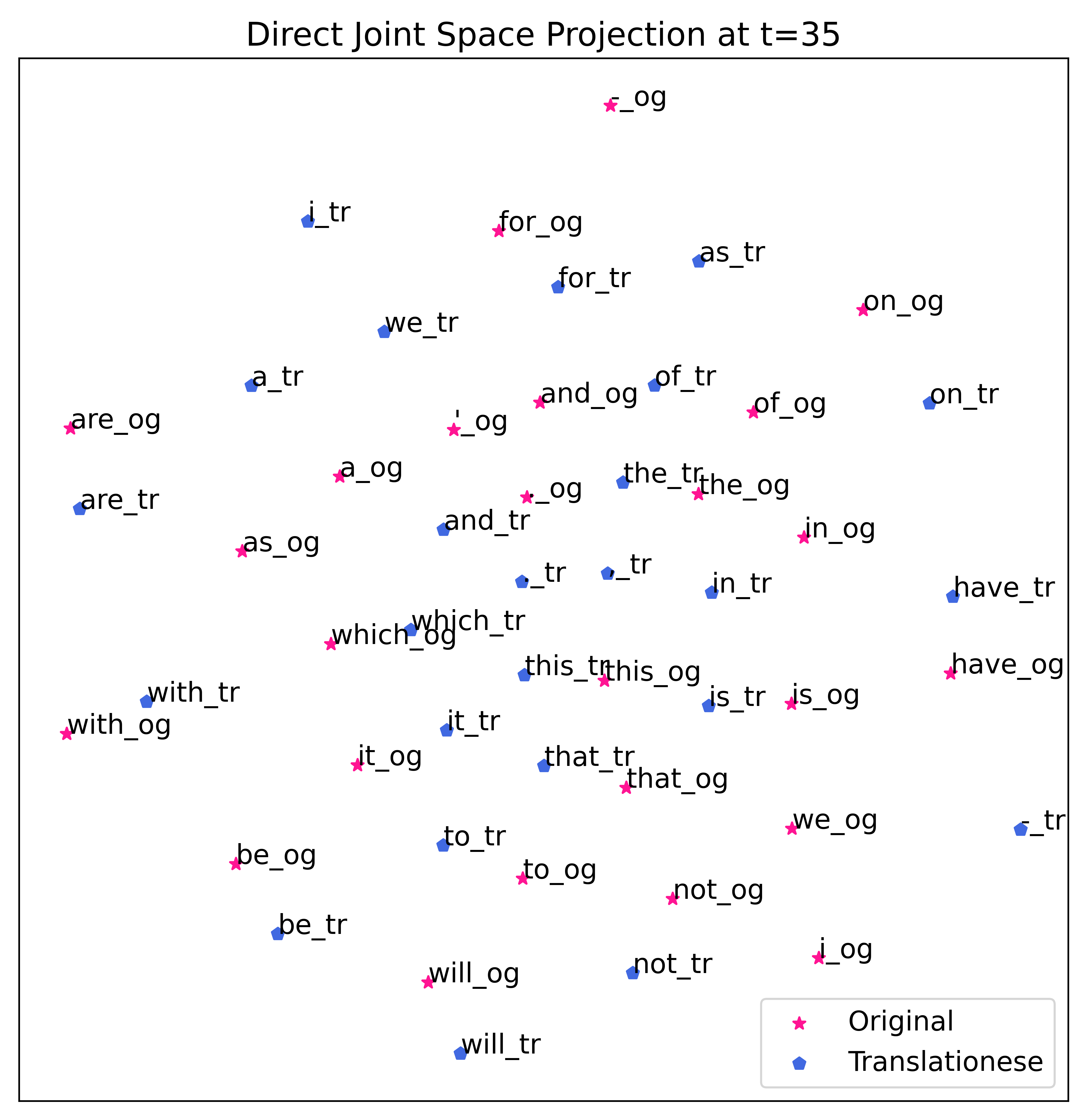}
\caption{Clustering on Direct Joint Space, before (up-side) and after (down-side) debiasing.}
\label{dendogram}
\end{figure}

Figure~\ref{dendogram} shows the t-SNE \citep{van2008visualizing} projection of the vectors in Direct Joint Space Projection before and after debiasing with INLP.

\subsection{Translationese Word Lists}
\label{app:wordlists}
The size of the translationese word lists created via the usage change algorithm of \citep{gonen-etal-2020-simple} is shown in Table~\ref{tab:wordlist-stats} and our top-50 elements per language are shown in Table~\ref{t:lists}.

\begin{table*}[th]
\centering
\resizebox{0.82\textwidth}{!}{%
\begin{tabular}{r cccccc}
\toprule
\textbf{rank}  &  De-En  &  De-Es  &  En-De  &  En-Es  &  Es-De  &  Es-En      \\
\midrule
1  &  hin  &  weise  &  each  &  hand  &  ahí  &  -      \\
2  &  heißt  &  art  &  while  &  down  &  ”  &  ”      \\
3  &  zumindest  &  gilt  &  hand  &  mind  &  -  &  modo      \\
4  &  hohe  &  hinaus  &  upon  &  labour  &  supuesto  &  duda      \\
5  &  dessen  &  sowohl  &  direct  &  while  &  terreno  &  supuesto      \\
6  &  dabei  &  interesse  &  mean  &  due  &  igual  &  algún      \\
7  &  außerdem  &  bleibt  &  secondly  &  form  &  “  &  bajo      \\
8  &  weise  &  tat  &  light  &  example  &  incluso  &  pues      \\
9  &  bleibt  &  außerdem  &  mind  &  others  &  luego  &  general      \\
10  &  völlig  &  gesamte  &  beyond  &  another  &  sí  &  igual      \\
11  &  derzeit  &  beim  &  takes  &  high  &  largo  &  cual      \\
12  &  sorge  &  ebenso  &  kind  &  each  &  allá  &  luego      \\
13  &  sowohl  &  beispiel  &  comes  &  food  &  origen  &  sí      \\
14  &  daran  &  kommt  &  practice  &  throughout  &  )  &  tipo      \\
15  &  darin  &  letztendlich  &  capacity  &  whole  &  bajo  &  cabo      \\
16  &  art  &  dabei  &  full  &  true  &  orden  &  alto      \\
17  &  seite  &  davon  &  itself  &  call  &  precisamente  &  tercer      \\
18  &  ihrem  &  sorge  &  open  &  single  &  corresponde  &  único      \\
19  &  beispielsweise  &  heißt  &  sort  &  air  &  solamente  &  resulta      \\
20  &  erster  &  linie  &  share  &  capacity  &  interés  &  fuera      \\
21  &  macht  &  weder  &  labour  &  organisation  &  general  &  incluso      \\
22  &  beispiel  &  völlig  &  large  &  power  &  idea  &  línea      \\
23  &  beim  &  zwar  &  words  &  practice  &  tercer  &  pública      \\
24  &  bzw.  &  erster  &  short  &  become  &  objeto  &  largo      \\
25  &  ebenso  &  genau  &  individual  &  takes  &  pública  &  siempre      \\
26  &  gilt  &  darin  &  nor  &  without  &  saben  &  alguna      \\
27  &  bedeutet  &  selbst  &  value  &  nature  &  donde  &  )      \\
28  &  hinaus  &  dessen  &  sense  &  reality  &  alto  &  términos      \\
29  &  voll  &  diejenigen  &  allow  &  comes  &  da  &  donde      \\
30  &  grund  &  jedem  &  least  &  doubt  &  algún  &  orden      \\
31  &  gesamte  &  macht  &  side  &  euro  &  cara  &  asimismo      \\
32  &  unser  &  ihrem  &  behind  &  close  &  línea  &  riesgo      \\
33  &  jedem  &  zumindest  &  currently  &  words  &  toda  &  ningún      \\
34  &  steht  &  eindeutig  &  longer  &  far  &  asimismo  &  “      \\
35  &  tatsache  &  sei  &  doubt  &  course  &  misma  &  mientras      \\
36  &  kommt  &  seite  &  whole  &  circumstances  &  propio  &  vista      \\
37  &  eindeutig  &  sorgen  &  down  &  currently  &  (  &  claro      \\
38  &  interesse  &  innerhalb  &  effect  &  non  &  través  &  efecto      \\
39  &  tatsächlich  &  form  &  become  &  yet  &  fuera  &  práctica      \\
40  &  all  &  handelt  &  outside  &  long  &  duda  &  régimen      \\
41  &  ganze  &  gesamten  &  indeed  &  itself  &  cual  &  ella      \\
42  &  hohen  &  ebene  &  board  &  least  &  central  &  misma      \\
43  &  weder  &  jeder  &  free  &  every  &  alguna  &  público      \\
44  &  aller  &  teil  &  needs  &  fisheries  &  resulta  &  carácter      \\
45  &  gleichzeitig  &  beispielsweise  &  without  &  interest  &  claro  &  plan      \\
46  &  gesamten  &  liegt  &  rural  &  general  &  junto  &  defensa      \\
47  &  allein  &  weit  &  throughout  &  main  &  tipo  &  dentro      \\
48  &  ort  &  schließlich  &  across  &  mean  &  demás  &  lado      \\
49  &  innerhalb  &  übrigen  &  close  &  free  &  falta  &  número      \\
50  &  jeder  &  reihe  &  another  &  term  &  siempre  &  personal      \\
\bottomrule
\end{tabular}
}
\caption{Top-50 translationese words as obtained by the application of the use change concept for the three languages L1-L2 (L1 being En, Es and De) when they are translated from the other two languages L2.  }
\label{t:lists}
\end{table*}

\end{document}